\documentclass[acmsmall]{acmart}

\AtBeginDocument{%
  \providecommand\BibTeX{{%
    \normalfont B\kern-0.5em{\scshape i\kern-0.25em b}\kern-0.8em\TeX}}}

\setcopyright{acmlicensed}
\acmJournal{PACMCGIT}
\acmYear{2024} \acmVolume{7} \acmNumber{2} \acmArticle{etra24-fp1065} \acmMonth{6}\acmDOI{10.1145/3654706}

\acmArticle{1}

\acmConference[ETRA '24]{Make sure to enter the correct
  conference title from your rights confirmation emai}{June 04--07,
  2024}{Glasgow, United Kingdom}
\usepackage{subfig}
\usepackage{pifont}
\usepackage{wrapfig}

\newcommand{\bfsection}[1]{\vspace*{0.1cm}\noindent\textbf{#1.}}
\newcommand{\itsection}[1]{\vspace*{0.1cm}\noindent\textit{#1.}}

\newcommand{\eg}{\textit{e.g.}}

\begin{document}

\title[Learning Gaze-aware Compositional GAN]{Learning Gaze-aware Compositional GAN from Limited Annotations}

\author{Nerea Aranjuelo}
\email{naranjuelo@vicomtech.org}
\orcid{0000-0002-7853-6708}
\email{naranjuelo@vicomtech.org}
\affiliation{%
  \institution{Fundación Vicomtech, Basque Research and Technology Alliance }
  \country{Spain}
}

\author{Siyu Huang}
\affiliation{%
  \institution{Clemson University}
  \country{USA}}
\email{siyuh@clemson.edu}
\orcid{0000-0002-2929-0115}

\author{Ignacio Arganda-Carreras}
\affiliation{%
  \institution{University of the Basque Country}
  \country{Spain}
}
\affiliation{%
  \institution{Ikerbasque}
  \country{Spain}
}
\affiliation{%
  \institution{Donostia International Physics Center}
  \country{Spain}
}
\affiliation{%
  \institution{Biofisika Institute}
  \country{Spain}
}
\email{ignacio.arganda@ehu.eus}
\orcid{0000-0003-0229-5722}

\author{Luis Unzueta}
\affiliation{%
 \institution{Fundación Vicomtech, Basque Research and Technology Alliance }
 \country{Spain}}
\email{lunzueta@vicomtech.org}
\orcid{0000-0001-5648-0910}

\author{Oihana Otaegui}
\affiliation{%
  \institution{Fundación Vicomtech, Basque Research and Technology Alliance}
  \country{Spain}}
\email{ootaegui@vicomtech.org}
\orcid{0000-0001-6069-8787}

\author{Hanspeter Pfister}
\affiliation{%
  \institution{Harvard John A. Paulson School of Engineering and Applied Sciences} 
  \state{MA}
  \country{USA}
  }
\email{pfister@g.harvard.edu}
\orcid{0000-0002-3620-2582}

\author{Donglai Wei}
\affiliation{%
  \institution{Boston College}
  \state{MA}
  \country{USA}}
\email{weidf@bc.edu}
\orcid{0000-0002-2329-5484}

\renewcommand{\shortauthors}{Aranjuelo N., et al.}

\begin{abstract}
Gaze-annotated facial data is crucial for training deep neural networks (DNNs) for gaze estimation. However, obtaining these data is labor-intensive and requires specialized equipment due to the challenge of accurately annotating the gaze direction of a subject.
In this work, we present a generative framework to create annotated gaze data by leveraging the benefits of labeled and unlabeled data sources.
We propose a Gaze-aware Compositional GAN that learns to generate annotated facial images from a limited labeled dataset. Then we transfer this model to an unlabeled data domain to take advantage of the diversity it provides. 
Experiments demonstrate our approach's effectiveness in generating within-domain image augmentations in the ETH-XGaze dataset and cross-domain augmentations in the CelebAMask-HQ dataset domain for gaze estimation DNN training. We also show additional applications of our work, which include facial image editing and gaze redirection.
\end{abstract}

\begin{CCSXML}
<ccs2012>
   <concept>
       <concept_id>10010147.10010257.10010293.10010294</concept_id>
       <concept_desc>Computing methodologies~Neural networks</concept_desc>
       <concept_significance>300</concept_significance>
       </concept>
   <concept>
       <concept_id>10003120.10003121.10003128</concept_id>
       <concept_desc>Human-centered computing~Interaction techniques</concept_desc>
       <concept_significance>100</concept_significance>
       </concept>
   <concept>
       <concept_id>10010147.10010371.10010382.10010383</concept_id>
       <concept_desc>Computing methodologies~Image processing</concept_desc>
       <concept_significance>500</concept_significance>
       </concept>
 </ccs2012>
\end{CCSXML}

\ccsdesc[300]{Computing methodologies~Neural networks}
\ccsdesc[100]{Human-centered computing~Interaction techniques}
\ccsdesc[500]{Computing methodologies~Image processing}
\keywords{Gaze estimation, synthetic data, GAN, generative, DNN, domain transfer}

\begin{teaserfigure}
\centering
  \includegraphics[width=0.76\textwidth]{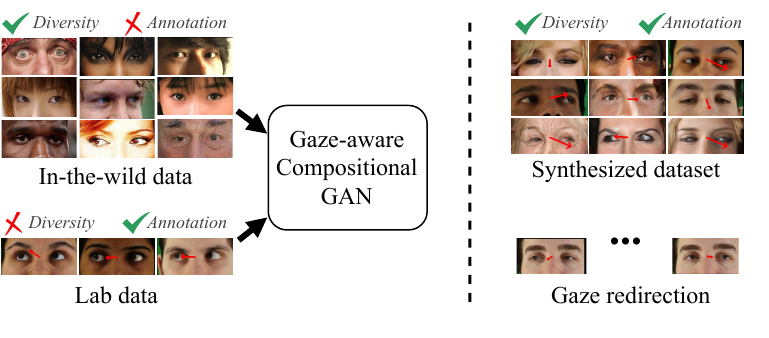}
  \vspace{-0.15in}
  \caption{Left: The proposed Gaze-aware Compositional GAN model can leverage the limited gaze annotation in the lab data and diverse appearance in the unlabeled data. Right: Applications include gaze-annotated image dataset generation and gaze redirection.}
  %\Description{Enjoying the baseball game from the third-base seats. Ichiro Suzuki preparing to bat.}
  \label{fig:samples} 
\end{teaserfigure}

\received{10 November 2023}
\received[revised]{February 2024}
\received[accepted]{March 2024}

\maketitle

\section{Introduction}

Gaze estimation is an essential task in computer vision for many applications, including but not limited to human-computer interaction, virtual reality, and the automotive industry. %For example, gaze estimation can be used to create more natural ways to interact with computers design more immersive experiences in virtual worlds, or improve driving safety by detecting driver distraction. 
Recent works in gaze estimation have demonstrated that Deep Neural Networks (DNNs) are more robust and accurate than traditional methods~\cite{cheng2021appearance}.
% challenge
However, obtaining enough quality data is essential for training DNNs for gaze estimation. 
%\begin{wrapfigure}{r}{0.55\textwidth}
%\includegraphics[width=\linewidth]{images/teaser_v2_tight2.pdf}
%\vspace{-0.3in}
%\caption{\label{fig:samples} Left: The proposed Gaze-aware Compositional GAN model can leverage the limited gaze annotation in the lab data and diverse appearance in the unlabeled data. Right: Applications include gaze-annotated image dataset generation and gaze redirection.}
%\end{wrapfigure}
Despite the abundance of facial images through the Internet and open large-scale datasets~\cite{lee2020maskgan}, obtaining gaze-annotated images remains a major challenge. Manually annotating existing images is tedious and prone to errors, as accurately determining where a person is looking is difficult.
Capturing gaze-annotated data is labor-intensive and typically requires specialized equipment and long capturing sessions, where a volunteer is asked to direct their gaze to various points ~\cite{zhang2020eth}.

Given the high value of each captured sample, it is crucial to explore alternative methods for obtaining them. One potential solution is to use synthetic data%, which has gained significant attention for training DNNs %due to the difficulty and effort involved in obtaining real-world training data 
~\cite{nikolenko2021synthetic}. 
Data might be generated using 3D simulated environments. However, this approach involves its own difficulties, such as the need for additional processes to bridge the gap between the simulated and real data domains. Synthetic data generation and augmentation using generative models might be a promising alternative. Generative adversarial networks (GAN) can produce high-quality images with a high level of realism ~\cite{karras2019style} with the added benefit of controlling the generated data by editing latent input vectors or conditioning the model on some input information~\cite{shu2021gan}.

In this paper, we present a Gaze-aware Compositional GAN learned from limited annotations. As shown in Fig.~\ref{fig:samples}, the proposed model can be used for both discriminative tasks like gaze estimation by synthesizing gaze-annotated facial images and generative tasks like gaze redirection. Our approach relies on a composition-based gaze-aware generative model to augment a limited annotated dataset.
Our method consists of two stages. First, we train a compositional GAN to generate gaze-aware images using a small annotated training dataset. Second, we transfer the model to a bigger in-the-wild unlabeled dataset. Then, we can generate within- and cross-domain new images to augment the limited annotated data. Our code is available at https://github.com/naranjuelo/GC-GAN. 
%allows to generate new samples from the same domain and from an additional unlabeled one. 
% for obtaining gaze data or for exploiting already existing data

The main contributions of this paper are:

% target task: gaze estimation from limited data
% DA
% synthetic dataset: datasetGAN can’t handle it
% solution: image editing with inversion latent/control, SSGAN ++
\begin{itemize}
    \item We introduce a novel approach for acquiring annotated gaze estimation data from limited annotations, leveraging the strengths of various labeled and unlabeled data sources. 
    \item We develop a Gaze-aware Compositional GAN that generates realistic synthetic images with a user-specified gaze direction using a limited annotated dataset. We transfer the gaze-aware generative model to an unlabeled dataset to exploit the variance of in-the-wild captured data.
    \item We validate our method by demonstrating its ability to generate previously unseen within-domain and cross-domain image augmentations that boost the accuracy of a gaze estimation DNN. We show our method can be used for different applications, such as data generation, image editing, data augmentation, and gaze redirection.
%    \item 
\end{itemize}

\section{Related work}
\label{sec:soa}

\subsection{Gaze Estimation from Facial Images} 
Traditional model-based methods for estimating gaze rely on a 3D eye model and use detected features such as the pupil center or the corneal reflections to estimate the gaze ~\cite{ishikawa2004passive}. Recent appearance-based approaches use DNNs to estimate the gaze using images as input ~\cite{zhang2020eth,abdelrahman2022l2cs,gazetr2022}. These approaches adapt better to challenging in-the-wild scenarios but strongly rely on large amounts of varied data for training. Public available datasets might alleviate this need. Some of these datasets are captured in controlled environments (e.g., a laboratory) and typically include relatively few subjects but a wide range of balanced gaze samples ~\cite{xia2020controllable,zhang2020eth}. Datasets obtained in more natural conditions typically include a wider variety of subjects but lack the controlled wide range of gazes ~\cite{zhang2015appearance,kellnhofer2019gaze360}.
Due to the variability in person-specific appearance and the influence of the setup (e.g., face perspective), gaze estimation DNNs are frequently fine-tuned before deployment with some samples captured in the target setup ~\cite{masko2017calibration,park2019few,arar2016regression}. Typically, this involves a tedious process where users sequentially look at different targets. Thus, DNNs are often retrained with a limited number of images. Due to the challenges of getting accurately gaze-annotated data, synthetic data emerge as an alternative to data capturing.

\subsection{Gaze-aware Facial Image Generation} 
 Realistic eye image synthesis has been extensively studied for its importance in various applications. There are two main approaches: 3D eye modeling and image generation using generative models. 

% synthetic data from 3D models
Graphics-based methods produce high-resolution images based on a 3D model of the human eye region%, which is used in a rendering framework
~\cite{wood2016learning}. Although gaze directions and head poses can be controlled in these approaches, using artificial eye texture can make the generated images appear unrealistic and widen the discrepancies between generated and target data, leading to lower accuracy when training DNNs with these unrealistic data. %~\cite{wood2018gazedirector} combines real images and a 3D eye region modeling, which is fitted and rendered for a specific gaze direction.
% GANs for refiment
Some works propose to reduce the domain discrepancies by leveraging GANs to enhance the realism of synthetic samples before using them for DNN training~\cite{shrivastava2017learning}. 
%SimGAN ~\cite{shrivastava2017learning} uses an adversarial network to refine synthetic data. %They use a pixel-level deviation penalty to maintain gaze direction, but this constraint reduces image diversity. 
GazeGAN~\cite{sela2017gazegan} employs unpaired image-to-image translation to transform images between domains.
 These methods involve generating synthetic data from 3D models, which can be time-consuming due to 3D asset design and rendering. %Furthermore, they require an additional refinement restricted to a single-eye region.

% NeRFs:
Some recent works use Neural Radiance Fields (NeRFs)~\cite{mildenhall2021nerf} to synthesize photoreal images with gaze control. In~\cite{li2022eyenerf}, mesh-based and volumetric reconstruction are combined to synthesize the eye region. Still, it lacks modeling of facial expression-induced eye deformations, requires a complex image capture setup, and struggles with details like eyelashes. %GazeNeRF~\cite{ruzzi2023gazenerf} proposes a two-stream architecture to predict volumetric features for the face and eye regions separately before compositing them. 
%NeRF-based works~\cite{li2022eyenerf,ruzzi2023gazenerf} focus on obtaining 3D-consistent results rather than capturing the diversity of images under a gaze-controlled setting and require many input images for accurate 3D mesh generation~\cite{remondino2023critical}.

To overcome the aforementioned challenges, our work generates realistic images from latent vectors using GANs. 
%with no need for prior graphics-based image generation.
The state-of-the-art GANs~\cite{brock2018large,karras2021alias} 
%, such as BigGAN ~\cite{brock2018large} and StyleGAN3 ~\cite{karras2021alias}, 
can generate realistic images without a 3D simulation environment. Our work is inspired by the recent composition-based generation SemanticStyleGAN ~\cite{shi2022semanticstylegan}. Although many of these GANs focus on facial image generation, gaze-aware data generation remains an unexplored challenge.

\subsection{Dataset Generation for Gaze Estimation} 

Given the limited and hard-to-obtain gaze data, GANs might be convenient to augment the existing data. Many works have attempted to use GANs to achieve controllable image generation or editing. %, frequently for face editing. 
This can be tackled by learning a model to manipulate the latent space of a pre-trained GAN ~\cite{shen2020interpreting,harkonen2020ganspace} or training with additional supervision to learn a more disentangled latent space ~\cite{deng2020disentangled,shi2022semanticstylegan}. These works are often applied to face attributes edition, such as hair or skin, but not gaze direction. Furthermore, most of the generated images tend to look forward as it is the most common gaze in typical training datasets ~\cite{lee2020maskgan,karras2019style}.
GANs also have limitations, based on what they have seen during training ~\cite{jahanian2019steerability}, so the training data domain influences their generation capabilities.% of the trained model to generate varied images. 

%\textbf{Gaze correction.}
Specific studies focus on gaze correction or redirection. %~\cite{he2019photo} proposes a cycle consistency and perceptual loss to redirect the gaze of input eye crops. 
InterpGaze ~\cite{xia2020controllable} generates intermediate gaze images given two samples using an encoder, a controller, and a decoder. ~\cite{chen2021coarse} also uses an encoder-decoder architecture, followed by a refinement generator. 
These works are limited to redirecting the gaze of a cropped eye region.
Some works also learn to blend this crop into the face, such as DeepWarp ~\cite{ganin2016deepwarp} and GazeGAN ~\cite{zhang2019gazecorrection}, but are constrained to gaze correction or limited movements. % <-- these 2 are compared in the experiments
EyeGAN~\cite{kaur2020eyegan} uses a segmentation mask instead of an RGB image to generate an image for a gaze direction. However, it is limited to tight grayscale eye-crop images.  
% do not take advantage of GANs for image and DBs generation (e.g., datagan)
Our approach is not limited to gaze redirection but allows gaze-aware image generation. Using a composition-based architecture, we can control specific face components for data augmentation. Finally, combining datasets from different domains and with different strengths enhances the variability of images our method generates.

\begin{figure*}[t]
\centering
\includegraphics[width=0.94\linewidth]{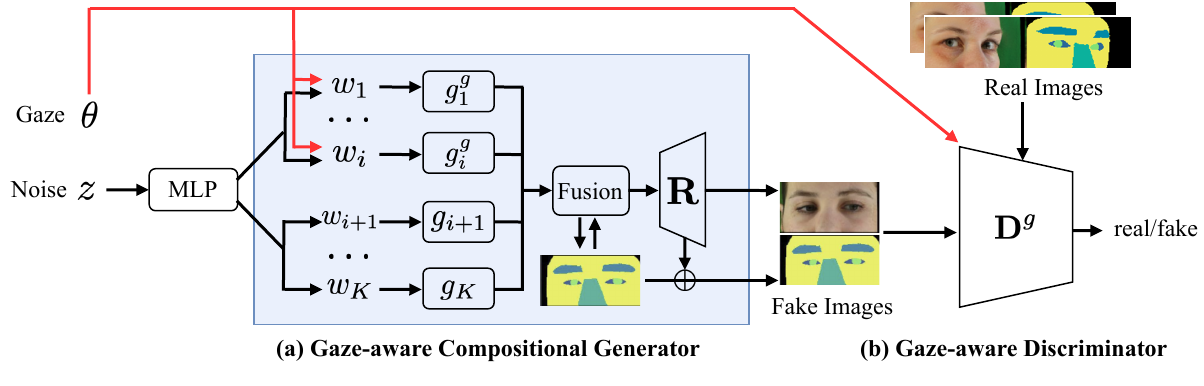} 
\vspace{-0.15in}
\caption{\label{fig:stage1} The proposed Gaze-aware Compositional GAN model (GC-GAN). Building upon the SemanticStyleGAN~\protect\cite{shi2022semanticstylegan}, we first group facial components into gaze-related $\{w_1,\dots,w_i\}$ and gaze-unrelated $\{w_{i+1},\dots,w_K\}$. We extend (a) the Local Generators ($\{g_j^g\}_1^i$) in the generator and the discriminator $D^g$ to condition on the input gaze $\theta$.}
\end{figure*}

\section{Method}

%3.1. Overview

%3.2. Stage-1: Conditional Gaze Generation

%put module details here

%3.3. Stage-2: Transfer to Unlabeled Domain

%3.4 Inference and Applications

%\subsection{Overview}
In a typical GAN, a generator model is trained to map a noise vector $z$ %(usually from a standard normal distribution) 
to an image $x$. Our proposed method learns to generate an eyes' region image $x$ and a corresponding segmentation mask $y$ given a vector $z$ and a target gaze direction $\theta$ the eyes should follow, as defined below: 
\begin{equation}
    G : (z, \theta) \longrightarrow (x, y) \,,
\end{equation}
where $y \in \{0,1\}^{HxWxK}$, being $H$ and $W$ the image dimensions, and $K$ the number of segmented face components. 
% The model learns to model the joint distribution of the image, the mask, and the gaze. 
% Note that having the same input vector $z$ and a slightly different $\theta'$ should produce the corresponding modification only in the pixels belonging to the eyes of the synthesized image, but not in the rest of the image.
Our method allows training on labeled and unlabeled domains, which adds the challenge of having no annotations to supervise the synthesized images' gaze in training.

\subsection{Gaze-aware Compositional GAN}
%\subsection{Stage 1: Training on Labeled Data}
\label{sec:stage1}

We have the following three observations:
(1) Some facial components are related to the gaze (\eg, iris), while others are not (\eg, nose).
(2) Our facial image generator needs to be compositional, as we need the gaze-related facial components to remain the same during the gaze-invariant data augmentation for the gaze estimation task and to be changed alone during the gaze redirection editing task.
(3) Our discriminator needs to penalize the sample if it doesn't match the conditioned gaze direction in addition to being not realistic. Thus, we designed the following model components.

%\bfsection{Framework: Gaze-aware SemanticStyleGAN}
\bfsection{Overall framework}%: Gaze-aware SemanticStyleGAN}
To ensure the composability of our GAN, we base our model on the state-of-the-art SemanticStyleGAN model~\cite{shi2022semanticstylegan}.
We extend SemanticStyleGAN to support effective gaze conditioning while generating realistic images and learning from labeled and unlabeled domains.
As illustrated in Fig.~\ref{fig:stage1}, our Gaze-aware Compositional GAN model (GC-GAN) is divided into three main components: noise to latent vectors mapping, Gaze-aware Compositional Generator for image generation, and Gaze-aware Discriminator for discriminating images during training. 
The input vector $z$ is mapped to an intermediate latent code $w$ $\sim$ $W$ using an 8-layer MLP to better model the non-linearity of the data distribution, similar to ~\cite{karras2019style}. To obtain a disentangled latent space for different face components, the latent code $w$ is divided into $K$ local latent codes and an additional base latent code $\mathbf{w}^{base}$, common for all the face components. 
These latent codes are processed by the Gaze-aware Compositional Generator. This module has $K$ generators, each responsible for generating feature maps for a specific face component given a local latent code $w_k$. All the output feature maps are fused and fed to the final generator, which generates synthetic images and the corresponding segmentation masks. During training, the generated images and masks are fed to the discriminator, as well as real samples from training data. Gaze vectors are also input to the discriminator when available. The main modules of the model are detailed as follows.

\begin{figure}%[t]
\centering
\includegraphics[width=0.95\textwidth]{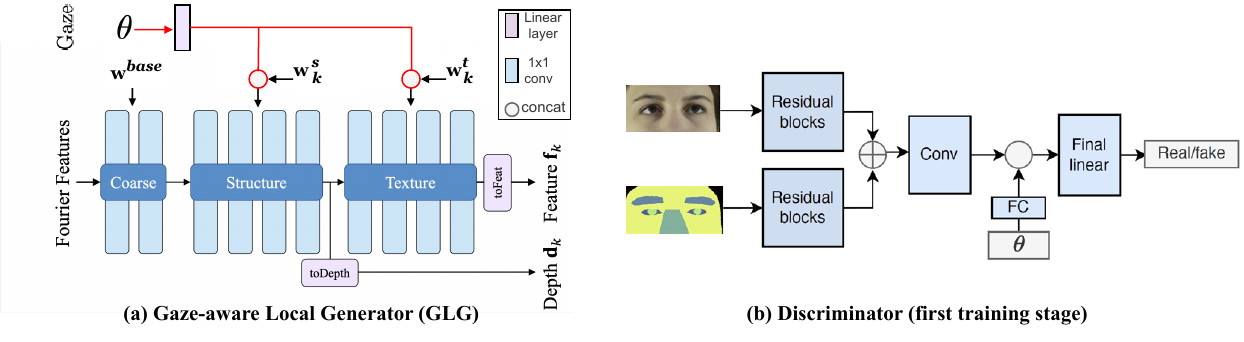}
\vspace{-0.15in}
\caption{\label{fig:archs} Architectures of the GC-GAN. Gaze-aware Local Generator (GLG) (a): the red lines are additions compared to~\cite{shi2022semanticstylegan}. The discriminator in the first training stage (b): outputs from residual blocks are summed, while features from gaze direction concatenated.}
\end{figure}

\bfsection{Gaze-aware generator and discriminator}
%\bfsection{Gaze-aware Compositional Generator} %{Local Generators.}
For image generation, as the Render Net is gaze agnostic, we only need to condition the gaze direction for the Local Generators of gaze-related facial components, $\{g_j^g\}_1^i$.
For gaze-related features, we condition the Local Generators with the gaze direction feature. We refer to these generators as Gaze-aware Local Generators (GLG).
Fig.~\ref{fig:archs} (left) shows the GLG architecture for a specific facial component $k$. 
Each local generator is formed by modulated $1\times1$ convolution layers with latent code-conditioned weights, and input Fourier features $f_p$ for position encoding. The input latent codes for each local generator are the base latent code $\mathbf{w}^{base}$, and the face component-specific $w_k$. The latent code $w_k$ is divided into shape and texture latent vectors, $w_{k}^{s}$ and $w_{k}^{t}$~\cite{shi2022semanticstylegan}.
The target gaze direction $\theta$ is also input to the GLGs so that we can control the gaze of generated samples. 
The input gaze $\theta$ is defined as yaw and pitch angles ($\theta_y$, $\theta_p$), which are fed to a fully connected layer for mapping them to an adequate 64-dimensional space before fusion. The output of this layer is concatenated to the component-specific latent vectors, and the extended latent vectors are fed to a series of modulated $1\times1$ convolutions and final fully connected layers.
The output of the GLG is a 1-channel pseudo-depth $d_{k}$ and 512-channel feature map $f_{k}$. The hidden layers have 64 channels. The local generators unrelated to gaze follow the same architecture except for the gaze branch, which we discard. 

%\begin{figure}[t]
%\centering
%\includegraphics[width=0.55\linewidth]{images/GLG.pdf}
%\caption{\label{fig:glg} Architecture of our Gaze-aware Local Generator (GLG). The red lines are additions compared to Shi \textit{et al.}~\protect\cite{shi2022semanticstylegan}.}
%\end{figure}

%\bfsection{Gaze-aware Discriminator}
Regarding the discriminator, our generated samples are discriminated based on the joint distribution of the image, the mask, and the gaze (Fig.~\ref{fig:archs}, right). For the image and mask, the discriminator has two convolution branches %composed of residual blocks 
whose outputs are summed up. Each branch is formed of residual blocks, similar to~\cite{karras2019style,shi2022semanticstylegan}, which convert the input $256\times256$ images to $4\times4$ feature maps. The mini-batch standard deviation of the summed feature maps is concatenated, and $3\times3$ kernel convolution is applied before reshaping to a 1-dimensional vector and adding the data from the gaze. The gaze is included as a third input to the discriminator. A linear layer maps the gaze to a 64-dimensional space before being concatenated with feature maps from the dual branch. These maps are processed by the final linear layer, classifying the input data as real or fake.

\begin{figure}[h]
\centering
\includegraphics[width=0.55\linewidth]{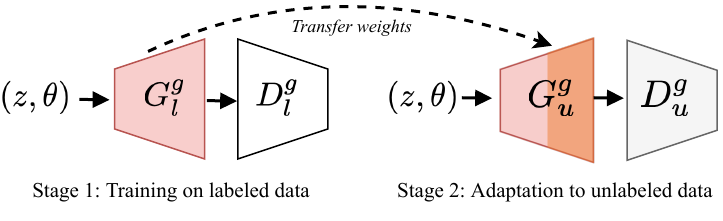}
\vspace{-0.13in}
\caption{\label{fig:pipeline} Two-stage training: (1) training on limited annotated data, (2) adaptation to unlabeled data.}
\end{figure}

\vspace{-0.1in}
%\subsection{Knowledge Transfer to Unlabeled Domain}
%\subsection{Stage 2: Adaptation to Unlabeled Domains}
\subsection{Two-stage Training}
%\subsection{Model transfer to unlabeled data}
\label{sec:transfer}
We first train the GC-GAN model on the labeled data and then transfer the facial appearance from the unlabeled data with the domain adaptation approach (Fig.~\ref{fig:pipeline}).

\bfsection{Stage 1: Training on labeled data}
During the first stage, the GC-GAN (Section \ref{sec:stage1}) is trained by minimizing the loss %function

\begin{equation}
\label{eq:losses_s1}
    L_{s1} = \lambda_{l}L_{l} + \lambda_{r}L_{r} + \lambda_{p}L_{p} + \lambda_{m}L_{m} + \lambda_{s}L_{s} \,,
\end{equation}
% D: logisticLoss(R/F) + R1 loss 
% G: non-saturating logistic loss + mask loss + segLoss + path regularization loss + path lossRegu
where $L_{l}$ is non-saturating logistic loss ~\cite{creswell2018generative}, $L_{r}$ is R1 regularization loss ~\cite{mescheder2018training}, $L_{p}$ is path length regularization loss ~\cite{karras2019style}, $L_{m}$ and $L_{s}$ are mask and segmentation regularization loss ~\cite{shi2022semanticstylegan}. Each loss is weighted with the corresponding $\lambda$ (Section \ref{sec:exps}).
However, due to the limited amount of labeled data, it is hard for such a trained model to generalize well to new in the wild images that are challenging to have accurate gaze annotations.

\bfsection{Stage 2: Adaptation to unlabeled data}
To adapt GANs to the target domain, it is a common practice to freeze the lower-level layers of the generator module~\cite{benaim2018one}. The challenge here is to find out which module to freeze and, additionally, how to handle the lack of annotations in the unlabeled domain.
In our case, the target domain lacks the label information that is used in the source domain, which requires the modification of the discriminator model.

\itsection{Freezing Gaze-aware Compositional Generator} To transfer the model to the appearance distribution from a new domain, we rely on the hypothesis that samples with the same segmentation mask share the same gaze direction. 
A specific latent vector $w$ and input gaze direction $\theta$, fed to the gaze-aware generators module, results in a fused coarse feature map and mask, which are then refined in the Render Net. Our goal is that the same latent vector generates images in both dataset domains that share the same gaze direction and coarse features (e.g., pose), but domain-specific appearances. %For that purpose, we first train the model with a labeled dataset.
For that purpose, we initialize the model with the pretrained weights from the first stage, freeze the GLG and the first block of the rendering generator, and fine-tune the rest of the model. Random input gazes are fed to the generator.

%\begin{wrapfigure}{r}{0.51\textwidth}
\begin{figure}[h]
\centering
\includegraphics[width=0.4\linewidth]{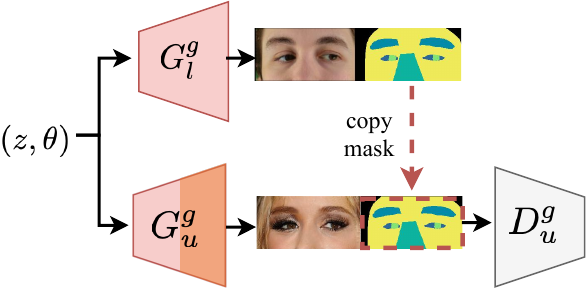}
\vspace{-0.2in}
\caption{\label{fig:stage2} Stage 2: Adaptation to unlabeled data. The generated image is combined with the mask generated by the pretrained stage-1 model given the same latent and gaze vectors.}
\end{figure}
%\end{wrapfigure}

%\vspace{-0.1in}
\itsection{Modifying Gaze-aware Discriminator}  There are no available gaze annotations to use as input along with real images and masks, so we constrain the gaze requirement by using the segmentation mask. %Instead of using the mask generated for each latent by the training model, we use the mask generated by the pretrained model for the specific latent. 
The discriminator has the same architecture as in Section \ref{sec:stage1} without the input gaze branch. 
The discriminator receives real pairs of RGB images and the corresponding segmentation masks, and the same with synthetic pairs. In the synthetic pairs, we use the synthetic image generated by the generator given a latent vector $w$ and a target gaze direction $\theta$, but an expected segmentation mask for that latent vector instead of the generated one.
The expected mask is generated by the previously pretrained model, as shown in Fig.~\ref{fig:stage2}. The model trained in the first stage
has already learned how to model the relation between the latent code and the gaze in the generated images. Consequently, using it to generate the masks for the discriminator's input pairs forces the learning GAN to generate images that fit the given masks and, hence, the input gaze direction.
%matching RGB images and masks, which are also aligned with the input gaze direction.

We also add a mask loss $L_g$, which is minimized together with the same losses as the first stage. This loss helps to preserve the same generated mask for the same latent vector and, thus, the same gaze direction. The new mask loss measures the mean squared error between the generated mask with the updated weights and the mask generated using the pretrained weights given the same latent and gaze vector. The final loss of the second stage is defined as follows:

\begin{equation}
    L_{s2} = L_{s1} + \lambda_{g}L_{g} \,,
    % [+loss: R1  + mask + stylegan2]
\end{equation}
where $L_{s1}$ are the losses defined in Equation ~\ref{eq:losses_s1} and $L_{g}$ is the mask loss for gaze preservation.

%\subsection{Inference and Applications}
\subsection{Applications}
%\subsection{Synthetic image editing} % (Inference)}
\label{sec:application}
\bfsection{Gaze-aware facial image synthesis}
The composition-based architecture promotes a disentangled and interpretable latent space for image editing. After model training, given an intermediate latent vector $w$ and a target gaze $\theta$ we can generate a synthetic sample. We can redirect the gaze of this sample by modifying the input gaze direction $\theta$. Regarding the semantic facial components, we can modify the specific $k_{th}$ component by varying the corresponding local latent vector ${w}_k^s$ or ${w}_k^t$.
We use cubic spline interpolation to smoothly move from a local latent vector to a different one.

\begin{figure}[h]
\centering
\includegraphics[width=0.88\linewidth]{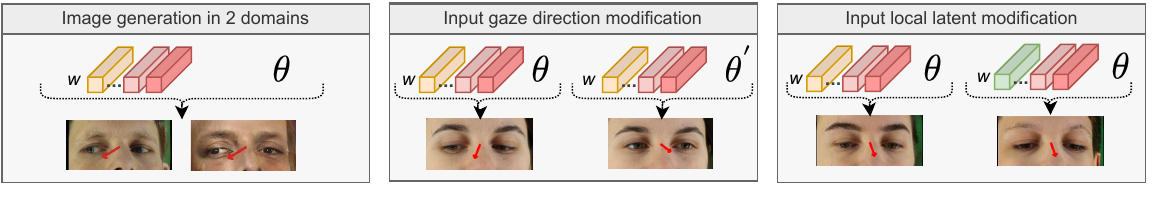}
\vspace{-0.2in}
\caption{\label{fig:augs_inf} Design choices for data augmentation with the trained GC-GAN model.}
\end{figure}

\bfsection{Data augmentation for gaze estimation}
Fig. \ref{fig:augs_inf} shows different ways for generating data we propose: data generation in different domains using the same latent and gaze vector, redirecting the gaze, and modifying specific facial components.

%\vspace{-0.05in}

\vspace{-0.1in}
\bfsection{Gaze-aware facial image editing}
To edit the gaze direction of a facial image, we need to map it to the learned latent space through image inversion. This process involves optimizing a latent vector to generate an image that closely resembles the desired target image. We use the loss:
\begin{equation}
\label{loss_inv}
L_{inv} = \lambda_p L_p + \lambda_d L_d + \lambda_m L_m \,,
\end{equation}
where the perceptual loss $L_p$ measures the high-level similarity between the generated and target images computing the L2 distance between intermediate image features using a pretrained VGG19 network ~\cite{zhang2018unreasonable}.
The $L_d$ computes the pixel-wise MSE loss between both images. Finally, the $L_m$ computes the difference between the evaluated latent vector and a mean latent vector to prevent significant deviations. The mean latent vector is computed by averaging multiple latent vectors from random $z$ vectors. All losses are weighted with the corresponding $\lambda$ (Section \ref{sec:exps}).

%The \verb|title|
%command has a ``short title'' parameter:
%\begin{verbatim}
 % \title[short title]{full title}
%\end{verbatim}

\section{Experiments}

\label{sec:exps}
%implementation details
We thoroughly examine our proposed framework through qualitative and quantitative experiments to showcase its effectiveness in generating data augmentations, as well as gaze-controlled images.

%\bfsection{Datasets} 
\subsection{Datasets and Implementation Details}
\bfsection{Datasets}
\label{sec:exp_details}
We use the ETH-XGaze dataset~\cite{zhang2020eth} as the primary source of gaze-labeled data. It contains images of 110 subjects captured from 18 cameras in a controlled environment.
% Data augmentation alleviates data scarcity in a target domain, 
We randomly select a subset of 8 subjects for training (14,464 images, subject IDs: $0$, $3$, $28$, $29$, $52$, $55$, $81$, $83$) and 2 for testing (3,920 images, subject IDs: $19$, $24$) for the four most frontal cameras (camera IDs: $1$, $2$, $3$, $8$). While the ETH-XGaze dataset features a wide and balanced gaze range, the variety and naturalness of individuals captured in controlled environments may be limited. To address this limitation, we use the CelebAMask-HQ dataset ~\cite{lee2020maskgan} as the unlabeled dataset, comprising 30,000 in-the-wild celebrity images.

\begin{figure}[h]
\centering
\includegraphics[width=0.68\linewidth]{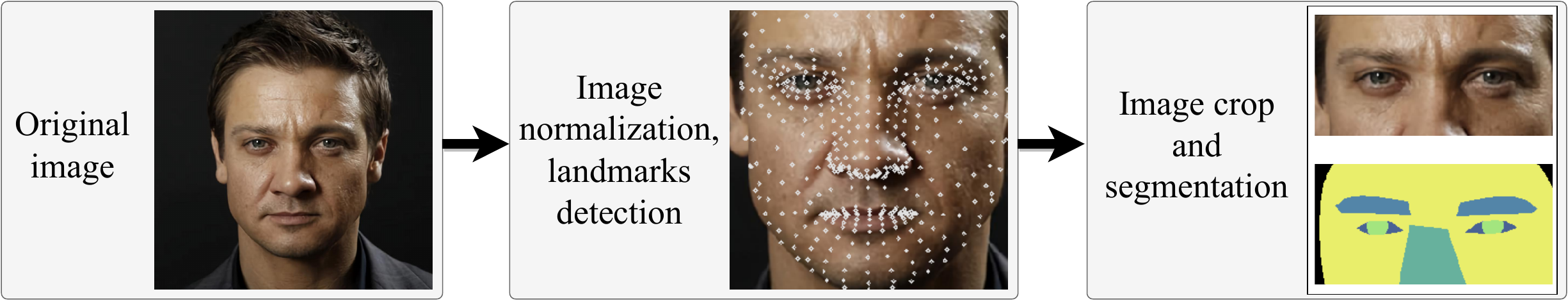}
\vspace{-0.1in}
\caption{\label{fig:prepr} Preprocessing of a sample from the CelebAMask-HQ dataset: face and face landmarks are detected for image normalization and eyes' region image crop and segmentation.}
\end{figure}
\vspace{-0.1in}
\bfsection{Data preprocessing}
We preprocess the data from the ETH-XGaze and CelebAMask-HQ datasets.
It is common to use image crops containing one or both eyes for gaze estimation DNNs ~\cite{shrivastava2017learning,masko2017calibration,porta2019u2eyes,sinha2021flame}. We use the eyes' region crop for our experiments. %Individual eyes can be cropped after synthetic image generation if needed. 
For models using the entire face, our approach can be extended to the face.
We preprocess the images to get a normalized $256\times256$ eyes' region image crop and the corresponding segmentation mask. 
We first detect the faces in the images~\cite{bazarevsky2019blazeface} and their face landmarks~\cite{kartynnik2019real}.
%use the face detector in ~\cite{bazarevsky2019blazeface} to detect the faces in images, and we detect the face landmarks with ~\cite{kartynnik2019real}. % to get the eyes' region crop and segment the target face components. 
We normalize each image by rotating it to eliminate any roll angle due to the head pose, centering it based on the face center, and scaling it to have 1.7 times the eyes' region width, considering the furthest eye corners horizontally. We define the face center as the average point considering the average of the eyes' coordinates and the mouth's coordinates. The eyes' region is the facial image's upper half. We use the face landmarks to generate the corresponding segmentation mask, which includes the following categories: background, face, iris, sclera, eyebrows, and nose. Fig.~\ref{fig:prepr} shows an example for a CelebAMask-HQ dataset sample.

\bfsection{Training} We first train the composition-based GAN with the ETH-X Gaze subset and then transfer the model in the second stage to the CelebAMask-HQ dataset.
We set $K$ to 6 for the considered number of components in the image. We have 4 non-conditioned local generators (for background, face, eyebrows and nose) and 2 GLGs (for sclera and iris).  
%SemanticStyleGAN implementation and maintain the training settings. 
Random input gaze angles are fed to the model during training, sampled from the real gaze range of the labeled data. We set a dimensionality of 512 for $z$ and $w$, and the image size is $256\times256$. %, as eyes images used in gaze estimation models usually have no higher resolution~\cite{shrivastava2017learning,masko2017calibration,porta2019u2eyes,sinha2021flame}.
Similar to StyleGAN2, we use style mixing regularization ~\cite{karras2019style} and leaky ReLU activations. We empirically set the first stage's $\lambda$ hyperparameters to $\lambda_{l}$ = 1, $\lambda_{r}$ = 10, $\lambda_{p}$ = 2, $\lambda_{m}$ = 100, and $\lambda_{s}$ = 500. In the second stage, $\lambda_{g}$ is set to 100. We train our model for 250k iterations in the first stage and 470k iterations in the second.
We use the ADAM optimizer ~\cite{kingma2014adam} with $\beta_1$ = 0, $\beta_2$ = 0.99, and 8 as the minibatch. 
We implement our framework using PyTorch 1.7 and we use an Nvidia GPU GeForce RTX 3090.

\subsection{Data Augmentation for Gaze Estimation}
\label{subsec:aug}

To assess the effectiveness of our method for data augmentation, we train a gaze estimation DNN with and without augmented data, respectively, and evaluate the DNN accuracy on the test set. 
% baseline augs (1):  clahe, shiftscale, blur, hueSatVal
% baseline augs (2): hueSatval, brighness, RGBshift
We consider four configurations: (1) a baseline configuration with common augmentations with geometric and color modifications (e.g., CLAHE, shift and scale perturbations), (2) a variation of the baseline configuration that only employs color modifications (e.g., brightness variations), (3) GAN-based within-domain augmentations, and (4) GAN-based within and cross-domain augmentations.

Regarding the gaze estimation DNN training, we train an off-the-shelf ResNet-50
~\cite{he2016deep} as in ~\cite{zhang2020eth}. We train the DNN on $224\times224$ eyes' region images to estimate gaze direction as yaw and pitch angles. We use the ADAM optimizer with a starting learning rate of 1e-3, a batch size of 50 and train for 25 epochs with learning rate decay every ten epochs. We select the best training epoch for testing based on the validation set ($15\%$ of the training set). We also train two other state-of-the-art DNNs to see the suitability of our synthetic data with different architectures. Specifically, we use the training configurations and implementations provided for the L2CS~\cite{abdelrahman2022l2cs} and the transformer-based GazeTR~\cite{gazetr2022}.

%\begin{table}[]
%\centering
%\caption{Error (degrees) in the test set when trained with different kinds of augmentations.}
%\resizebox{0.65\linewidth}{!}{%
%\begin{tabular}{ll}\hline
% \textbf{Augmentation} &  \textbf{Error}$\downarrow$ \\\hline
%No Aug & 4.54 \\ 
%Color Aug & 4.47 \\ 
%Color+Geometry Aug & 4.52 \\ \hline
%Ours (in-domain) & 4.22 \\ 
%Ours (in- and cross-domain) & \textbf{3.86} \\ \hline
%\end{tabular}%
%}
%\label{tab:exp1_error}
%\end{table}

For GAN-generated augmented samples, we first embed the ETH-XGaze subset samples into the closest latent vectors. New samples are then generated through iterative modification of unconditioned face components and redirecting the gaze vector of some of the samples (Section \ref{sec:application}). We generate 8,000 synthetic images for the within-domain augmentations and further augment the training set by an additional 5\% with the cross-domain augmentations. We use the same inverted images' latent vectors to generate the synthetic images in the CelebAMask-HQ dataset domain. 
% For the baseline augmentations, we apply them randomly to the images during training.
%We generate 5 sets with an incremental number of synthetic images (from 2,000 images to 14,000). Fig.~\ref{fig:aug_ethx} shows an example of real image from the ETH-XGaze dataset and some synthetic images obtained by modifying specific components of the corresponding latent vector.

To evaluate the accuracy of the gaze predictions, we compute the gaze angular error (º) as:

\begin{equation}
    L_{gaze} = \frac{{g}\cdot{g'}}{{||g||}\ {||g'||}} \,
\end{equation}

where $g \in \mathcal{R}^3$ is the ground-truth gaze and $g' \in \mathcal{R}^3$ the predicted gaze vector. The trained DNNs predict pitch and yaw angles, which are converted to 3D vectors (no roll angle~\cite{zhang2015appearance}).

Table~\ref{tab:exp1_error} shows the error obtained by the DNN in the test set when trained with the different augmented sample configurations, as well as when trained without augmentations. 
While all augmentations lead to an error drop, the reduction is significantly higher when using GAN-based augmentations. Additionally, we observe that even though cross-domain augmentations belong to a different domain, they help to improve the model's generalization capability.

\begin{table}[h] 
%\begin{subtable}[c]{0.5\textwidth}
\centering
\caption{Error (º) in the ETH-XGaze test set when trained with limited data and different kinds of augmentations and DNN architectures~\cite{zhang2020eth,abdelrahman2022l2cs,gazetr2022}.}
%\resizebox{0.85\linewidth}{!}{%
\vspace{-0.1in}
\begin{tabular}{c|ccc|ccc}%\hline
 \textbf{Augmentation}  & \multicolumn{3}{c|}{\textbf{Number of images}}& \multicolumn{3}{c}{\textbf{Error $\downarrow$}}\\ 
    & ETH-X (r)& ETH-X (s)&CELEB (s)&  ResNet& L2CS& GazeTR\\\hline
  
No Aug  & 14,464& 0&0& 4.54 & 4.07 & 5.11\\ %\hline  
Color Aug  & 14,464& 0&0& 4.47 & 3.67 & 4.89\\ %\hline  
Color+Geometry Aug  & 14,464& 0&0& 4.52 & 3.39 & 4.81\\ \hline
Ours (in-domain)  & 14,464& 8,000&0& 4.22 & 3.16 & 5.04\\ 
Ours (in- and cross-domain)  & 14,464& 8,000&1,123& \textbf{3.86} & \textbf{3.13} & \textbf{4.65}\\ %\hline \hline
\end{tabular}%
%}
\label{tab:exp1_error}
\end{table}

\begin{figure}%[t]
%\centering
\includegraphics[width=0.94\linewidth]{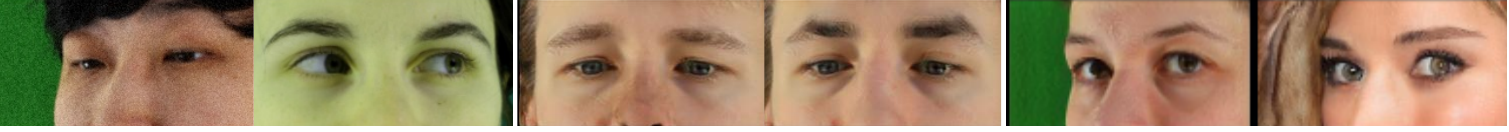}
\vspace{-0.1in}
\caption{\label{fig:aug_bothdoms} Different augmentations evaluated for DNN training. Left to right: color and geometric augmentations, within-domain augmentations, and cross-domain augmentations.}
\end{figure}
 
% [qualitative] augmented images (1 Figure)
%	= comparison with baseline images
%\vspace{-0.1in}
Fig.~\ref{fig:aug_bothdoms} shows qualitative results of the different augmentations, as evaluated in Table \ref{tab:exp1_error}. The images on the left have undergone geometry and color modifications, as used in the baseline augmentations configuration. The second pair of images are within-domain augmentations for the same subject. %, including the original image and modification of nose, eyebrows, and gaze direction. 
The images on the right depict a pair of synthetic images in the ETH-XGaze and CelebAMask-HQ data domains, generated using the same latent vector. It can be observed that despite differing appearances, images within each pair maintain consistent gaze direction.

%\vspace{-0.1in}
\subsubsection{Ablation Studies:} \textit{data generation for gaze estimation DNN training}

%4.3 Ablation studies
%	= Disc
%		# number of training data

%We present diverse ablation studies regarding data generation for gaze estimation DNN training.% and generative model training.
 
\begin{figure}[h]%[!tbp]
  \centering
  \subfloat[]{\includegraphics[width=0.48\textwidth]{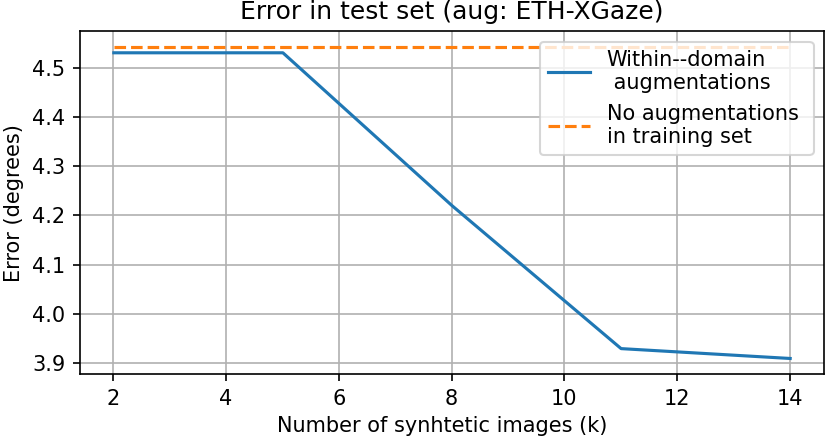}\label{fig:etest}}
  \hfill
  \subfloat[%Flower two.
  ]{\includegraphics[width=0.48\textwidth]{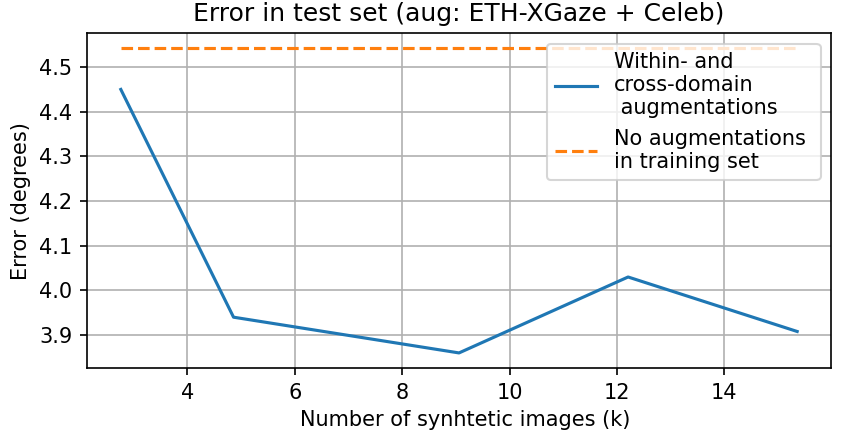}\label{fig:etest_cel}}
  \vspace{-0.13in}
  \caption{Mean error of the gaze estimation DNN (test set) when trained with varying number of synthetic samples from the ETH-XGaze data distribution (a) and the ETH-XGaze and CelebAMask-HQ data distribution(b).}
\end{figure}

%\subsection{Quantitative evaluation of data augmentation}
\bfsection{Data augmentation: in-domain synthesis}
We evaluate how varying the number of GAN-generated augmentations affects the accuracy of the gaze estimation DNN. We repeat the experiment in Section \ref{subsec:aug} with the DNN from~\cite{zhang2020eth}, varying the number of synthetic samples in the training set. Fig.~\ref{fig:etest} shows the obtained mean error in the estimated gaze for the different training sets adding within-domain augmentations.
When training with no synthetic samples, the DNN obtains an error of 4.53º in the test set. When we add some synthetic samples to the training set, the error is similar (4.52º), but as we increase the number of synthetic samples (more than 5,000 images), the error decreases to 3.90º. The error drop shows that the augmentations are helpful.% and the gaze is accurately preserved.

\bfsection{Data augmentation: cross-domain synthesis}
We increase each training set a $5\%$ by selecting the images with the highest confidence. We compute this confidence score based on the MSE between the generated masks in both dataset domains. 
We repeat the DNN training with different sets of augmented data.
Fig.~\ref{fig:etest_cel} shows the results in the test set when adding within-dataset and cross-dataset augmentations to the training set. It can be seen that adding cross-dataset augmentations helps in reducing the error. Adding images from a new domain increases the accuracy of the DNN, but adding an excessive amount of images does not seem to necessarily lead to further improvement (the minimum test error is 3.86º, and it is achieved with 9,000 synthetic images).

%\begin{figure}
%\centering
%\includegraphics[width=0.8\linewidth]{images/g2_resc.png}
%\caption{\label{fig:etest_cel} The mean error in the test set when using a gaze estimation DNN trained with a different number of synthetic samples from the ETH-XGaze and CelebAMask-HQ dataset distribution.}
%\end{figure}

\subsection{Gaze-aware Facial Image Generation}

%- [quantitative]
%	= baseline
%	= FID

\bfsection{Facial image synthesis} 
The quality of image synthesis in generative models is evaluated using metrics such as the Fréchet Inception Distance (FID) ~\cite{heusel2017gans} and the Inception Score (IS) ~\cite{salimans2016improved}. 
%\begin{wraptable}{r}{0.46\textwidth}
%\centering
%\caption{Image quality comparison with state-of-the-art methods for in-the-wild images' gaze redirection.}
%\vspace{-0.1in}
%\label{tab:quality}
%\begin{tabular}{lll}
%\hline
%\textbf{Method} & \textbf{FID} $\downarrow$ & \textbf{IS}$\downarrow$ \\ \hline
%DeepWarp                                  & 106.53                                 & 2.89                                  \\ 
%GazeGAN                                    & 30.21                                  & 3.10                                  \\\hline
%Ours                                      & \textbf{15.30}                                   & \textbf{1.92} %                                 \\ \hline
%\end{tabular}%
%}
%\end{table}
%\end{wraptable}
Our models obtain a mean FID of 15.3, and a mean IS of 1.92. As far as we know, there are no generative models with input latent vector and gaze to compare. Consequently, we compare these metrics to the state-of-the-art methods for in-the-wild images' gaze editing. DeepWarp and GazeGAN achieve a higher IS (2.89 and 3.10, respectively), while they also present a much higher FID (106.53 and 30.21, respectively). This suggests that our method can generate more realistic synthetic images. In addition, existing gaze redirection methods are typically limited to frontal gaze or gaze between two images, lacking fine-grained control.
%Table \ref{tab:quality} shows that while both DeepWarp and GazeGAN achieve a higher IS, they also present a much higher FID. 

\begin{figure}[htbp]
  \centering
  \begin{minipage}[b]{0.62\linewidth}
    %\centering
    \begin{tabular}{l|c}
        \textbf{Training data} & \textbf{Error $\downarrow$} \\ \hline
        ETH-XGaze & 4.88 \\
        ETH-XGaze + ~\cite{wood2016learning} & 4.70 \\
        ETH-XGaze + GC-GAN (in-domain) & 4.53 \\
        ETH-XGaze + GC-GAN (in-, cross-domain) & 4.51 \\
    \end{tabular}
    \vspace{-0.15in}
    \caption*{(a)}
  \end{minipage}%
  \begin{minipage}[b]{0.37\linewidth}
    %\centering
    \includegraphics[width=\linewidth]{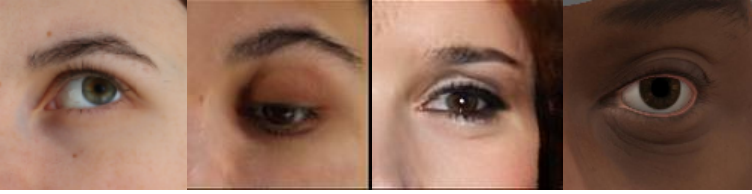}
    \vspace{-0.15in}
    \caption*{(b)}    
  \end{minipage}   
  \vspace{-0.12in}
  \captionof{figure}{\label{fig:eyecrop_results}Error in the ETH-XGaze test set (a) for single-eye crops with images generated by a non-DL method compared to GC-GAN. Samples of each training dataset (b), from left to right: ETH-XGaze, GC-GAN generated (in-domain), GC-GAN generated (cross-domain), and 3D graphics-based generated~\cite{wood2016learning}.}
\end{figure}

As explained in Section~\ref{sec:soa}, instead of training a generative model, we could synthesize facial images using non-DL approaches, such as graphics-based methods. % allow for generating images with specific requirements. 
However, the potential domain gap between real and synthetic images may limit their effectiveness in training. We generate $8,000$ synthetic images using the 3D graphics-based method in~\cite{wood2016learning} to evaluate gaze estimation accuracy when trained with these images compared to GC-GAN-generated ones. We train the gaze estimation DNN from~\cite{zhang2020eth} with the same ETH-XGaze training set as the experiments in Tab.~\ref{tab:exp1_error}: only ETH-XGaze samples, in-domain and cross-domain synthetic images (as in Table~\ref{tab:exp1_error}), and additionally images generated using 3D graphics. 
\cite{wood2016learning} can produce images for a single eye, so we adjust our images by cropping the right half (left eye) to ensure equitable training conditions for the DNN. 
%We train the DNN with only ETH-XGaze samples, in-domain and cross-domain synthetic images (as in Tab.~\ref{tab:exp1_error}), and images generated using 3D graphics. 
Fig.~\ref{fig:eyecrop_results}a shows the obtained results in the test set.

\vspace{-0.1in}
The error when training only with original samples (4.88º) is reduced when we add synthetic samples. However, this drop is higher with images generated with GC-GAN (4.51º). The results suggest that even though all synthetic images help, GC-GAN-generated ones are more effective. In addition, our approach learns from a specific data distribution and does not require a 3D environment, which can be time-consuming due to 3D asset design and rendering.
%It can also be noticed that 
Errors are higher compared to Table\ref{tab:exp1_error} (image containing both eyes), probably because of the reduced information.

\begin{figure}[h]
\centering
\includegraphics[width=0.3\linewidth]{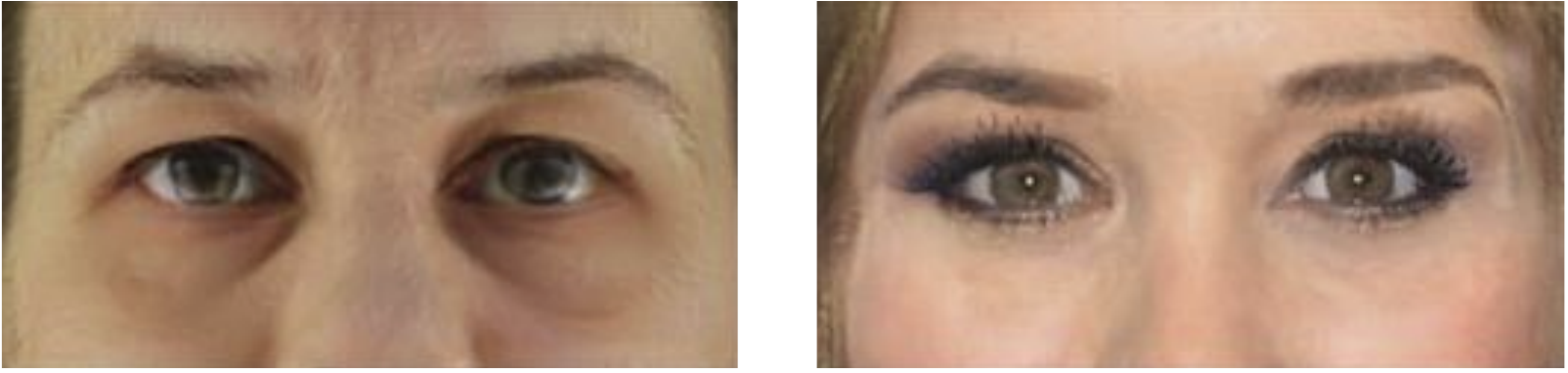}
\vspace{-0.1in}
\caption{\label{fig:both_sample_ad} Images generated by models of stage 1 and stage 2 given the same gaze $\theta$ and latent vector $w$.}
\end{figure}

\bfsection{Facial image editing} %Fig.~\ref{fig:face_edit} illustrates a synthetic image after redirecting the gaze (left) and after modifying the latent code for the facial component of the eyebrows  (right). It can be seen that specific face components can be edited without modifying the rest of the face. It can be seen, that our method allows editing different semantic facial parts without training any additional model to manipulate the latent space of the trained model, which makes the model more interpretable for the user. 
%\begin{figure}[h]
%\includegraphics[width=0.98\linewidth]{images/face_edit_cr_min.pdf}
%\caption{\label{fig:face_edit} Face edition of synthetic samples: gaze redirection (lefts) and eyebrows modifications (right).}
%\end{figure}
%- [qualitative] (1 Figure)
%	= gaze change
%	= facial component change: I1, I1+face-change, I1+eye-change
%	= appearance change: w/ celeb, w/o celeb
%Fig.~\ref{fig:both_modifs} displays examples of synthetic images generated from the same initial random latent vector in both dataset domains. It demonstrates the ability to selectively change different components of the face, such as the nose, eyebrows, and face shape while leaving the rest of the image unchanged by modifying the corresponding local latent vectors.
%\begin{figure}[h]
%\includegraphics[width=0.98\linewidth]{images/modifs_both_min2.pdf}%modifs_both_cr.pdf}
%\caption{\label{fig:both_modifs}Synthetic images generated from the same latent vector for ETH-XGaze and CelebAMask-HQ domains. Local latent vectors' modifications result in face components' modifications (left to right: eyebrows, nose, face shape).}
%\end{figure}
%\subsection{Image Editing in Different Domains}
The GC-GAN allows selectively changing different face components, such as the nose, eyebrows, and face shape, while leaving the rest of the image unchanged by modifying the corresponding local latent vectors.
%It can be seen that specific face components can be edited without modifying the rest of the face. 
Different semantic facial parts can be edited without training an additional model to manipulate the latent space of the model, which makes the model more interpretable for the user. 
%We also present some additional qualitative results on image editing in both domains. 
In Fig.~\ref{fig:both_sample_ad}, we see an example of two synthetic images generated with the models trained in the first and second stage, given the same latent vector $w$ and gaze direction $\theta$.

%\begin{wrapfigure}{r}{0.46\textwidth}
%\end{wrapfigure}

\begin{figure}[h]
\includegraphics[width=0.87\linewidth]{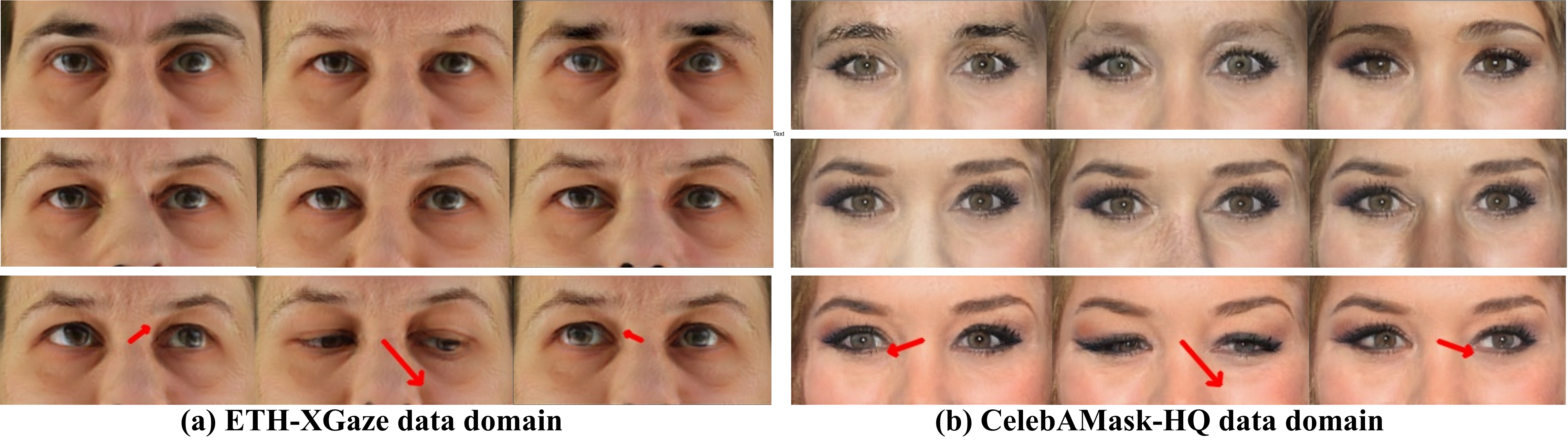}
\vspace{-0.12in}
\caption{\label{fig:augs_combi} Synthetic image augmentations by input latent or gaze vectors' modification in different domains. }
\end{figure}

%\vspace{-0.1in}
Next, we generate new augmented samples by editing a local latent vector $w_k$ or the input gaze direction $\theta$ (Section \ref{sec:application}
) of the shown samples. Fig.~\ref{fig:augs_combi} shows some examples generated by the model trained in the first and second stages, left and right images, respectively. %Fig.~\ref{fig:augs_celeb_ad} shows examples of the same kinds of augmentations generated with the model trained in the second stage (CelebAMask-HQ data domain).
%\begin{figure}[h]
%\includegraphics[width=0.65\linewidth]{images/exp_ethx.pdf}
%\caption{\label{fig:augs_ethx_ad} Synthetic image augmentations by input latent or gaze vectors' modification (ETH-XGaze data domain). }
%\end{figure}
We generate the images in the first row by modifying the latent vector of the eyebrows. We observe that augmented samples vary on the eyebrows but preserve the other face components unaltered, including the gaze direction. We generate the second-row images' by sampling new local latent vectors for the nose generator and leaving the rest of the inputs the same as in Fig.~\ref{fig:both_sample_ad}. %The third row shows examples of face shape modifications. 
The images in the last row are generated by modifying the input gaze $\theta$. 
%\begin{figure}[h]
%\centering
%\includegraphics[width=0.65\linewidth]{images/exp_celeb.pdf}
%\caption{\label{fig:augs_celeb_ad} Synthetic image augmentations by input latent or gaze vectors' modification (CelebAMask-HQ data domain).}
%\end{figure}
%As shown in Fig.~\ref{fig:augs_combi}, the proposed GC-GAN model allows fine-grained control on image editing. In addition to image editing, it is also crucial when generating training data for DNNs. 
Training datasets for DNNs should have enough samples and be varied and balanced. Consequently, having control over data generation is beneficial. 

\begin{figure}[h]
\centering
\includegraphics[width=0.87\linewidth]{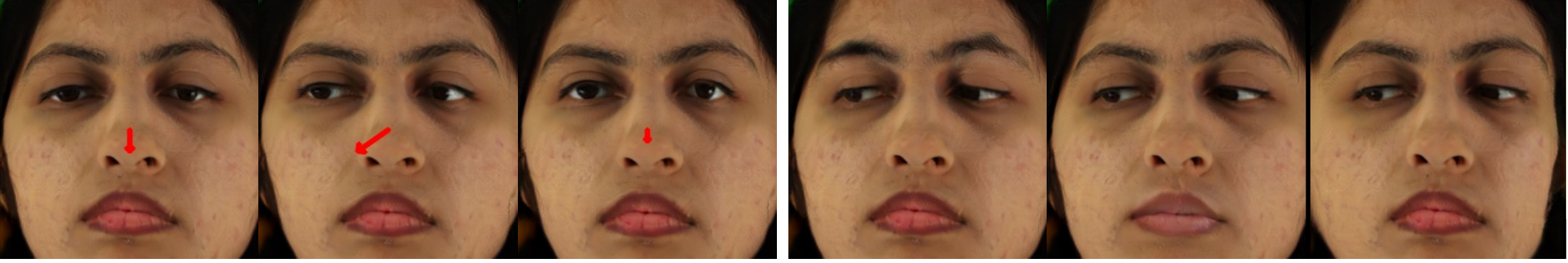}
\vspace{-0.1in}
\caption{\label{fig:faces_ethx} Synthetic image augmentations by input latent or gaze vectors' modifications when using complete face crops.}
\end{figure}

The focus of our method is to generate eye region images. However, our method can easily extend to complete faces in contrast to works such as ~\cite{shrivastava2017learning,kaur2020eyegan}. These images then could be used for gaze estimation DNNs that use complete face images. 
We train GC-GAN with complete face crops from the same limited subset of Section~\ref{sec:exp_details} and add the mouth as an additional component. The model obtains an FID of 27.1 and an IS of 1.91. 
Figure~\ref{fig:faces_ethx} shows qualitative results when training with complete faces.% and varying the input gaze (left) and different face components (right).

%% examples of face data generation

\begin{figure}[h]
\centering
\begin{minipage}{0.6\textwidth}
  \begin{tabular}{c|c|c|c|c||c}
    \textbf{Camera ID} & \textbf{1} & \textbf{2} & \textbf{3} & \textbf{4} & \textbf{Average} \\ \hline
    \textbf{Mean error (º)} & 3.80 & 2.94 & 5.85 & 7.73 & 5.08
  \end{tabular}
  \vspace{-0.1in}
    \caption*{(a)}
\end{minipage}%
\begin{minipage}{0.4\textwidth}
  %\centering
  \includegraphics[width=0.97\linewidth]{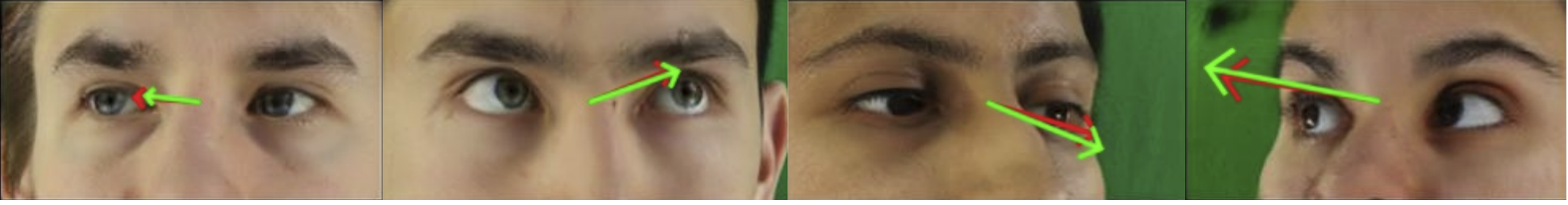}
  \vspace{-0.1in}
  \caption*{(b)}
\end{minipage}
\vspace{-0.2in}
\caption{Gaze accuracy of synthetic images: (a) error between estimated and input gaze direction to the GC-GAN and (b) target gaze (green) for image generation and estimated gaze (red) on the generated images for different cameras (from left to right, camera 1 to 4).}
\label{fig:gaze_acc_combi}
\end{figure}

\bfsection{Gaze direction accuracy} We measure image quality by the IS and FID metrics, but these metrics do not consider the correctness of the subject's gaze in the synthetic image. This gaze should be consistent with the input target gaze to the model. In Section~\ref{subsec:aug}, we show that the gaze is accurate enough to help in DNN training. Next, we evaluate the gaze accuracy explicitly. 
We compute the gaze error between the input gaze to the model and the gaze estimated by a gaze estimation DNN on the synthetic sample. To minimize the possible errors of the gaze estimation DNN, we train the models in the Section~\ref{subsec:aug} ~\cite{zhang2020eth,gazetr2022,abdelrahman2022l2cs} with the complete ETH-XGaze dataset and keep the two best checkpoints of each one. Then, we compute the mean of their output as the final gaze estimation for each tested image. We test all the synthetic images generated as image augmentations in Section~\ref{subsec:aug} (in- and cross-domain). Fig~\ref{fig:gaze_acc_combi}a shows the obtained mean error in degrees for each of the four camera perspectives included in the dataset.

%\begin{wraptable}{r}{0.46\textwidth}
%\caption{Error (º) between estimated gaze direction by an ensemble of DNNs and the input gaze direction to the GC-GAN.}
%\vspace{-0.1in}
%\centering
%\begin{tabular}{c|c|c|c|c||c}
%\textbf{Camera ID} & \textbf{1} & \textbf{2} & \textbf{3} & \textbf{4} & \textbf{Average} \\ \hline
%\textbf{Mean error (º)} & 3.80 & 2.94 & 5.85 & 7.73 & 5.08
%\end{tabular}%
%\label{tab:gaze_acc}
%\end{wraptable}

The mean error varies with perspective. More frontal perspectives (cameras 1 and 2) present lower error. Even if the obtained error is small, controlled-gaze image generation in lateral perspectives seems more challenging for the model. Fig.~\ref{fig:gaze_acc_combi}b shows some examples of synthetic images with arrows representing the gaze direction. Green represents the input target gaze to the GC-GAN, and red the estimated gaze by the DNN ensemble on the generated image. Despite some differences in gaze (from left to right, gaze errors: 3.33º, 1.78º, 4.7º, 8.84º), gaze directions are very similar.

%% examples of face data generation
%\begin{figure}[h]
%\centering
%\includegraphics[width=0.65\linewidth]{images/gaze_error.pdf}
%\vspace{-0.1in}
%\caption{\label{fig:gaze_error} Target gaze (green) for image generation and estimated gaze (red) on the generated images for different perspectives (from left to right, camera 1 to 4).}
%\end{figure}

\begin{figure}[h]
\centering
\includegraphics[width=0.98\linewidth]{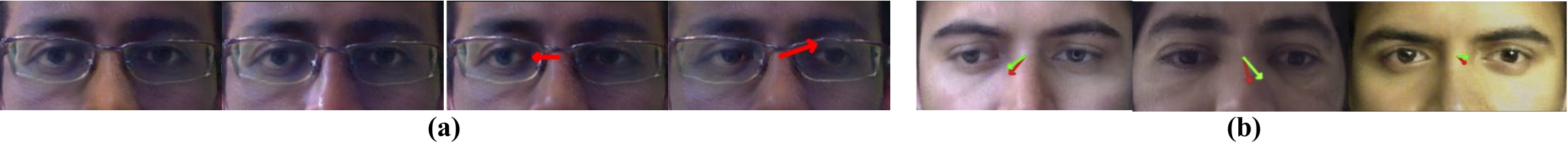}
%{images/mpii_ims_2_v3_.pdf}
\vspace{-0.15in}
\caption{\label{fig:gaze_error_mpii} Image augmentations by input latent or gaze vector modification in the MPIIFaceGaze data domain (a). Gaze accuracy in generated images (b): target gaze (green) for image generation and estimated gaze (red) on the generated images by the GC-GAN.}
\end{figure}

%\vspace{-0.15in}
\bfsection{Gaze direction accuracy in other benchmarks} To evaluate gaze accuracy under less controlled conditions compared to the ETH-X Gaze dataset, we extend the experiment to the MPIIFaceGaze dataset~\cite{zhang2015appearance}. Using the same training setup, we train the GC-GAN with data from 13 subjects in this dataset, achieving an FID of 17.66 and an IS of 1.84. As with the ETH-X Gaze augmented dataset (Section~\ref{subsec:aug}), we invert the original images to generate an augmented dataset, which we use to infer the gaze. For the gaze estimation DNN, we train the DNNs ensemble with all the images in the MPIIFaceGaze dataset and infer the gaze of generated 20,000 images. We obtain an average error of 5.29º, similar to Fig~\ref{fig:gaze_acc_combi}a. In this case, images are not captured in a laboratory and grouped by cameras. However, we obtain no significant difference between both dataset domains.
Fig.~\ref{fig:gaze_error_mpii}a shows examples of a synthetic image in the MPIIFaceGaze data domain with different augmentations (nose and gaze modifications).
Fig.~\ref{fig:gaze_error_mpii}b compares input target gaze to the GC-GAN (green) and estimated gaze directions by the DNNs ensemble (red) for some generated samples.

%\begin{wrapfigure}{r}{0.5\textwidth}
%\end{wrapfigure}

%\vspace{-0.1in}
\subsection{Ablation Studies: GC-GAN model training}

%\bfsection{Stage 2: Modules to freeze in the Gaze-aware Compositional Generator} 
%To evaluate the influence of different components of the method for knowledge transfer to the unlabeled domain, we conduct the following experiments. 
%We perform different ablation studies regarding GC-GAN model training.

%\begin{wraptable}{r}{0.5\textwidth}

\noindent
\textbf{What modules in Gaze-aware Compositional Generator should freeze during Stage-2?}
The generator has several combinations of modules to be frozen. We evaluate gaze transfer when training with different frozen modules. We qualitatively analyze gaze preservation by generating multiple pairs of images in both domains given a latent vector. 

\begin{table}[h]
\centering
\caption{Gaze preservation between domains when different components are frozen, Gaze-aware Local Generator ($GLG$) or Render Net ($R$), and whether mask constraint is applied.}
\vspace{-0.1in}
\label{tab:ablation_fr}
%\resizebox{\textwidth}{!}{%
\begin{tabular}{c|cc|c|c} \hline
\textbf{ID} & \textbf{GLG} & \textbf{R} & \textbf{$Mask_{c}$} &\textbf{Gaze preservation} \\  \hline
1   & shape only   & \ding{55}   &  \ding{51}         & No                         \\ 
2  &  \ding{51}      &  \ding{55}   &  \ding{51}        & No                         \\
3   & \ding{51}      & {1}   &  \ding{51}   & No                         \\ 
4   & \ding{51}      & {2}    &  \ding{51}  & Yes                        \\ 
%5   & \ding{51}      & {3}   &  \ding{51}   & Yes                      \\ 
5   & \ding{51}      & {2}   &  \ding{55}   & No   \\ \hline
\end{tabular}%
%}
\end{table}
%\end{wraptable}

\vspace{-0.1in}
In Table \ref{tab:ablation_fr}, we observe that freezing only the shape-oriented layers in the GLG is not enough, the weights of the whole GLG need to be preserved. The Render Net $R$ also contains some information related to the gaze, as it is required to freeze the 2 initial residual blocks to transfer the gaze. It is also possible to freeze an additional block, but the image quality is worse. Regarding the segmentation mask constraint (Section \ref{sec:transfer}), it is a necessary but not sufficient condition.

%We also experimented with the co-training of both stages, performing a single-stage training with no success. These experiments resulted in poor image quality and gaze control.

\bfsection{Components in Gaze-aware Generators}
The GC-GAN %groups generators for gaze-related and gaze-unrelated facial components using a composition-based architecture. This 
design enables independent control of components and a disentangled latent space. We rely on the hypothesis that if some facial components (e.g., nose) are gaze-independent, their generator should learn without gaze conditioning. To demonstrate this, we retrain the GC-GAN with the ETH-XGaze dataset but modify the Local Generators so that all of them are gaze-conditioned and have the same architecture as a GLG.

\begin{figure}%[h]
\includegraphics[width=0.92\linewidth]{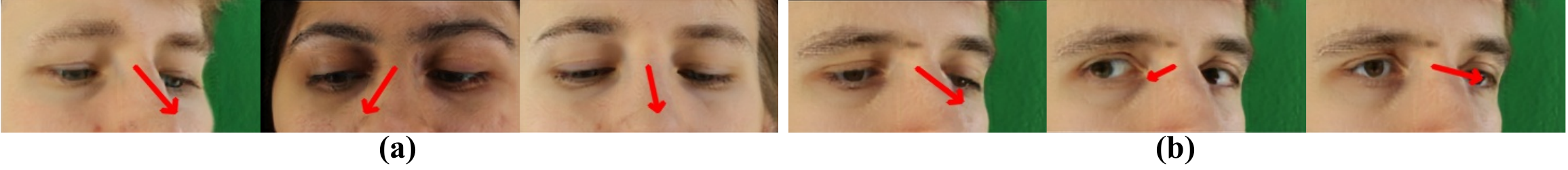}
\vspace{-0.18in}
\caption{\label{fig:conditioning} Synthetic images generated by varying the input gaze when all Local Generators are gaze-conditioned (a) and when generators are grouped as in GC-GAN (b).}
\end{figure}

We generate 5 random images with GC-GAN and the fully conditioned model. Then, we vary the input gaze of each sample to generate new images with 32 different gazes. We analyze how the images change, expecting only minor differences as only the eyes should vary. Differences are quantified using the pixel-wise mean absolute error between the original and redirected images. %We calculate the mean error for both models, %(Table~\ref{tab:ablation_cond}).
 Separating gaze-related and unrelated generators reduces the error (0.93 versus 1.33). Fig.~\ref{fig:conditioning} shows a sample when varying the gaze. When conditioning all the generators (a), the gaze is entangled with the pose. % - when gaze changes, so does pose. 
 The images generated with grouped generators (b) show a disentangled behavior: the eyes' appearance changes, but the rest of the image is unaltered.

\subsection{Applications for gaze estimation}

Our approach offers different potential applications to improve gaze estimation models or procedures. 
First, GC-GAN allows for combining labeled and unlabeled data sources, enabling training with unannotated in-the-wild images. This is particularly valuable given gaze annotations' laborious and costly nature and the importance of data quantity and variability in training DNNs. 
Our method can also be used to enhance the predictability of gaze estimation models by extensive model evaluation under different conditions. The synthetic images generated with GC-GAN could be used to evaluate a model's robustness against image variations and detect potential limitations or biases when using specific input data (e.g., different facial attributes, gaze distributions, or conditions, like glasses). This analysis could be used to improve models, predict their performance under different circumstances, and make their behavior more trustworthy. 
Additionally, our approach streamlines calibration by generating new samples within a domain and subject using a smaller image set. %, reducing the time and effort involved.
This enhances capturing efficiency and may improve model personalization (e.g., specific subject).

% - make models more explainable, transparent or reliable. Evaluate model behavior/robustness against different variations -> extend to other tasks(e.g., glasses) or detect possible biases to different face 
% - reduce calibration processes by generating new samples for the subjects

\bfsection{Privacy and ethics}
%In our study, we use publicly available datasets of facial images. 
The proposed method may inherit the negative societal impacts of existing deep generative models, such as improperly generating fake data. However, we believe our proposed method can be beneficial to build more robust gaze estimation models that guarantee universal human-computer interaction systems and have a smaller bias to certain people groups for correct behavior. Different approaches might be used to account for possible model biases or to avoid improper use. Variational autoencoders might be used to identify possible dataset bias~\cite{amini2019uncovering}, which synthetic images could later mitigate. Another possibility to promote responsible artificial intelligence is the data and model cards~\cite{pushkarna2022data,mitchell2019model}. Model cards are designed to accompany models to clarify their intended use, provide a model evaluation in various conditions (e.g., race, geographic location), or register the provenance of training datasets. 

\section{Conclusion}

In this paper, we have presented a novel method to generate annotated gaze data by combining the strengths of different labeled and unlabeled data sources. First, we train a gaze-aware compositional GAN that is able to generate realistic synthetic images with specific gaze directions in a labeled data domain. Then, we transfer this model to an unlabeled data domain to exploit the variance these data provide. Experiments have shown that our method can be used to augment existing annotated data for gaze estimation DNN training by generating within-domain and cross-domain augmentations that boost the DNN accuracy. 
The presented method does not require 3D virtual environments to generate synthetic data for DNN training and generates highly realistic samples, as the distribution of the generated images is learned from the training domain.
We have also shown other applications of our work to facial image editing and gaze redirection.
We believe the current work presents an exciting direction to leverage existing data and improve state-of-the-art computer vision models from a data-centric perspective.
In future work, we will explore incorporating the camera pose for fine-grained controlled image generation.

% when text is anonymous this part is not shown
\begin{acks}
This work has been partially funded by the Basque Government under the project AutoTrust (Elkartek-2023 program), the University of the Basque Country UPV/EHU grant GIU19/027, NSF-IIS grant 2239688 and NSF grant III-2107328. 
\end{acks}

\bibliographystyle{ACM-Reference-Format}
\bibliography{sample-base}

\appendix

%\section{Appendix title 1...}

\end{document}